\documentclass[10pt,twocolumn,letterpaper]{article}

\usepackage{cvpr}      % To produce the REVIEW version
\usepackage{algorithm}
\usepackage{algpseudocode}
\usepackage{amsmath} % 用于数学公式，如 \lfloor

\definecolor{cvprblue}{rgb}{0.21,0.49,0.74}
\definecolor{deepred}{RGB}{180, 0, 0}
\definecolor{orange}{RGB}{255, 165, 0}
\definecolor{deepcyan}{RGB}{35,140,137}
\usepackage[
    pagebackref,
    breaklinks,
    colorlinks,
    citecolor=orange,    
    linkcolor=deepcyan,      
    urlcolor=deepcyan,       
    filecolor=deepcyan       
]{hyperref}

\usepackage[most]{tcolorbox}
\usepackage{booktabs}
\usepackage{makecell}
\usepackage{graphicx} % for \resizebox
\usepackage{siunitx}
\usepackage{multirow}
\usepackage[table]{xcolor} 
\usepackage{caption}

\def\paperID{***} % *** Enter the Paper ID here
\def\confName{CVPR}
\def\confYear{2026}

\title{Enhancing Spatial Understanding in Image Generation via Reward Modeling}

\author{Zhenyu Tang\textsuperscript{\normalfont 1,2*} \quad
Chaoran Feng\textsuperscript{\normalfont 1*} \quad
Yufan Deng\textsuperscript{\normalfont 1,2} \quad
Jie Wu\textsuperscript{\normalfont 2} \quad
Xiaojie Li\textsuperscript{\normalfont 2} \quad \\ 
Rui Wang\textsuperscript{\normalfont 2} \quad  
Yunpeng Chen\textsuperscript{\normalfont 2} \quad
Daquan Zhou\textsuperscript{\normalfont 1} \\
\textsuperscript{\normalfont 1}\normalfont Peking University \quad
\textsuperscript{\normalfont 2}\normalfont ByteDance Seed
}
\begin{document}
\addtocontents{toc}{\protect\setcounter{tocdepth}{-1}}
\maketitle
{
  \renewcommand{\thefootnote}%
    {\fnsymbol{footnote}}
  \footnotetext[1]{Equal Contribution.}
}
\begin{abstract}
% Recent advances in image generation have been accompanied by stricter demands on prompts, which often input multiple complex spatial relationships and typically require repeated sampling to obtain satisfactory results. 
Recent progress in text-to-image generation has greatly advanced visual fidelity and creativity, but it has also imposed higher demands on prompt complexity—particularly in encoding intricate spatial relationships. In such cases, achieving satisfactory results often requires multiple sampling attempts.
To address this challenge, we introduce a novel method that strengthens the spatial understanding of current image generation models. We first construct the \textbf{SpatialReward-Dataset} with over 80k preference pairs. 
Building on this dataset, we build \textbf{SpatialScore}, a reward model designed to evaluate the accuracy of spatial relationships in text-to-image generation, achieving performance that even surpasses leading proprietary models on spatial evaluation.
% Through rigorous data curation and filtering, \textbf{SpatialScore} surpasses several leading vision-language models (VLMs) and existing reward models on spatial evaluation. 
We further demonstrate that this reward model effectively enables online reinforcement learning for the complex spatial generation. Extensive experiments across multiple benchmarks show that our specialized reward model yields significant and consistent gains in spatial understanding for image generation. 
The visual demo is available at the \href{https://dagroup-pku.github.io/SpatialT2I/}{project page}.
% All models and datasets will be released.
\end{abstract}    
\section{Introduction}
\label{sec:intro}

\begin{figure}[!ht]
    \centering
    \includegraphics[width=\linewidth]{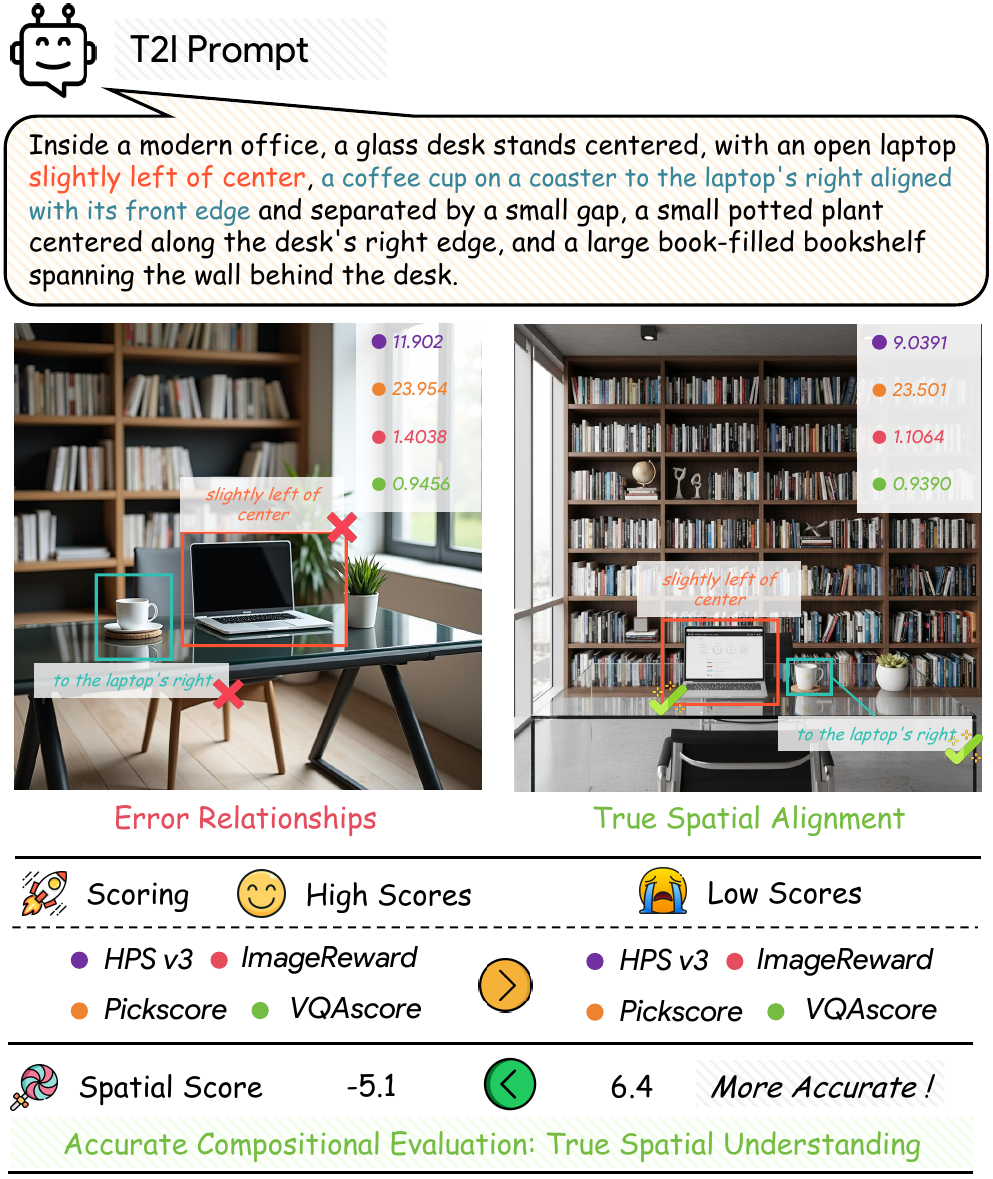}
    \vspace{-15pt}
    \caption{
    \textbf{Failure of Reward Models on Spatial Understanding.} 
    % Existing reward models~\cite{ma2025hpsv3,kirstain2023pickscore,xu2023imagereward,lin2024vqascore} incorrectly favor aesthetically pleasing but spatially inaccurate images (Left) over compositionally correct ones (Right). The higher scores for images with clear layout errors demonstrate that these models lack the fine-grained spatial reasoning required for robust T2I guidance.
    Existing reward models~\cite{ma2025hpsv3,kirstain2023pickscore,xu2023imagereward,lin2024vqascore} often assign higher reward values to spatially incorrect images than to spatially correct ones, thereby exposing their limited spatial reasoning capabilities.
    }
    \vspace{-15pt}
    \label{fig:duck_placeholder}
\end{figure}

Recent advances in generative models~\cite{ho2020denoising, song2020score,lipman2022flow-matching, liu2022flow-matching-rectified-flow, lin2024open, zhang2025epona, tang2025cycle3d, zhang2024repaint123,  peebles2023scalable} have transformed visual content creation, enabling the synthesis of high-fidelity and diverse images~\cite{rombach2022high, podell2023sdxl, flux2024,cao2025hunyuanimage, wu2025qwen-image, gao2025seedream}. Following the success of online reinforcement learning (RL)~\cite{shao2024deepseekmath} in large language models~\cite{guo2025deepseek-r1, openai2025}, recent studies~\cite{liu2025flow-grpo, xue2025dancegrpo, li2025mixgrpo, he2025tempflow-grpo, wang2025pref-grpo, wang2025grpoguard,zhou2025text} have explored applying GRPO-style reinforcement learning to diffusion models, leading to significant performance gains.

\begin{figure*}[!t]
    \centering
    \includegraphics[width=\linewidth]{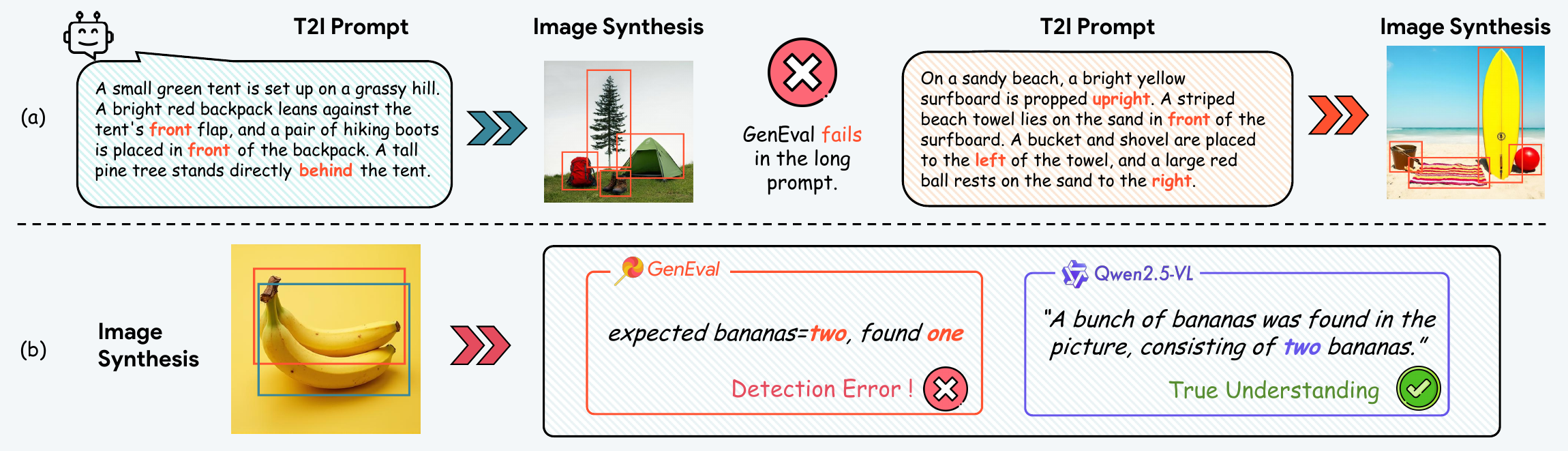}
    \vspace{-15pt}
    \caption{
    Limitations of  GenEval~\cite{ghosh2023geneval} as the reward model.
(a) GenEval-based RL training fails to generalize to long prompts involving complex spatial relationships across multiple objects.
(b) The rule-based GenEval rewards, which rely on object detectors, often produce incorrect evaluations under visual challenges like occlusion, while modern VLMs can accurately infer the correct response.
    % A comparison of evaluation paradigms: reasoning-based VLMs and score-based metrics. 
    % While score-based metrics like GenEval~\cite{ghosh2023geneval} fail to assess (a) complex spatial composition and (b) basic object counts, VLMs (Qwen2.5-VL~\cite{bai2025qwen2.5vl}) demonstrate the \textit{true understanding} required for these nuanced tasks. This highlights the necessity of the VLM-based approach that our work adopts.
    } 
    % \vspace{-15pt}
    \label{fig:geneval}
\end{figure*}

% \begin{figure*}[t!]
%     \centering
%     \includegraphics[width=\linewidth, height=4.5cm]{example-image-duck}
    
%     \caption{A placeholder figure using example-image-duck.}
%     \label{fig:duck_placeholder}
% \end{figure*}

However, with increasing prompt complexity, text-to-image models often struggle to accurately depict scenes that involve complex spatial relationships among multiple objects. This motivates us to explore how to enhance the spatial understanding of image generation models. Reinforcement learning emerges as a promising direction to address this challenge. However, despite its theoretical potential, applying online RL to improve spatial understanding remains difficult—primarily due to the lack of a reliable and effective reward model.

A straightforward approach is to adopt existing image reward models. As shown in Figure~\ref{fig:duck_placeholder}, human-preference reward models~\cite{xu2023imagereward,kirstain2023pickscore, wu2023hpsv2,ma2025hpsv3, wang2025unified} which incorporate text–image alignment as one of the evaluation factors, fail to accurately evaluate complex spatial relationships, and reward models~\cite{lin2024vqascore, hu2023tifa} designed for text-image alignment, which rely on VQA-style evaluations, exhibit the same limitation. The second option is to leverage the latest proprietary APIs of visual language models(VLM)~\cite{openai2025gpt5, comanici2025gemini}. However, their high cost makes them impractical for online RL, which requires frequent reward queries. The third alternative is to utilize open-source VLMs. However, our experiments show that even advanced models such as Qwen2.5-VL-72B~\cite{bai2025qwen2.5vl} suffer from substantial hallucinations and fail to provide reliable and accurate rewards, possibly because they are not optimized for complex reasoning on spatial relationships across multiple objects.

Recent works such as Flow-GRPO~\cite{liu2025flow-grpo} have also explored compositional  generation on the rule-based GenEval benchmark~\cite{ghosh2023geneval}, which computes rewards with an object detector and a color classifier. However, GenEval includes only simple prompts of the form \verb|"a photo of A <relative position> B."|
In Figure~\ref{fig:geneval}, we observe that training on GenEval fails to generalize to longer prompts with multiple spatial relationships, and its reward computation is highly sensitive to visual factors such as occlusion, leading to inaccurate rewards.

In this work, we argue that enhancing spatial understanding in image generation through online RL relies on constructing a reliable and accurate reward model. We first introduce  \textsc{SpatialReward-Dataset}, which contains 80K adversarial preference pairs spanning a wide range of real-world scenarios. Each preference pair is collected through an adversarial setup consisting of one image that accurately aligns with complex spatial relationships described in the prompt and one perturbed image that violates part of these relationships. To ensure data quality and accuracy, all pairs are carefully reviewed and filtered by human experts.

Building on this dataset, we further train \textsc{SpatialScore}, a powerful reward model designed to evaluate the accuracy of spatial relationships in image generation. Our results show that SpatialScore even surpasses several leading proprietary models, which exhibit hallucinations when reasoning about complex spatial relationships among multiple objects.

We further employ SpatialScore as the reward model for online RL. To improve training efficiency, we propose a top-$k$ filtering strategy that maintains a balanced sampling ratio between high-reward and low-reward candidates. The results on multiple benchmarks show that our approach effectively leverages feedback from SpatialScore, leading to substantial performance improvements over its base model.

Our contributions are summarized as follows:
\begin{itemize}
\item We introduce \textsc{SpatialReward-Dataset} with over 80K adversarial preference pairs, carefully curated by humans to ensure data quality.
\item We develop \textsc{SpatialScore}, a strong reward model for evaluating spatial relationship accuracy in image generation, which surpasses several leading proprietary models.

\item We employ SpatialScore as the reward model for online RL with a top-$k$ filtering strategy. Extensive experiments show substantial improvements in spatial understanding for image generation over the base model.

\end{itemize}

\begin{figure*}[t!]
    \centering
    \includegraphics[width=\linewidth]{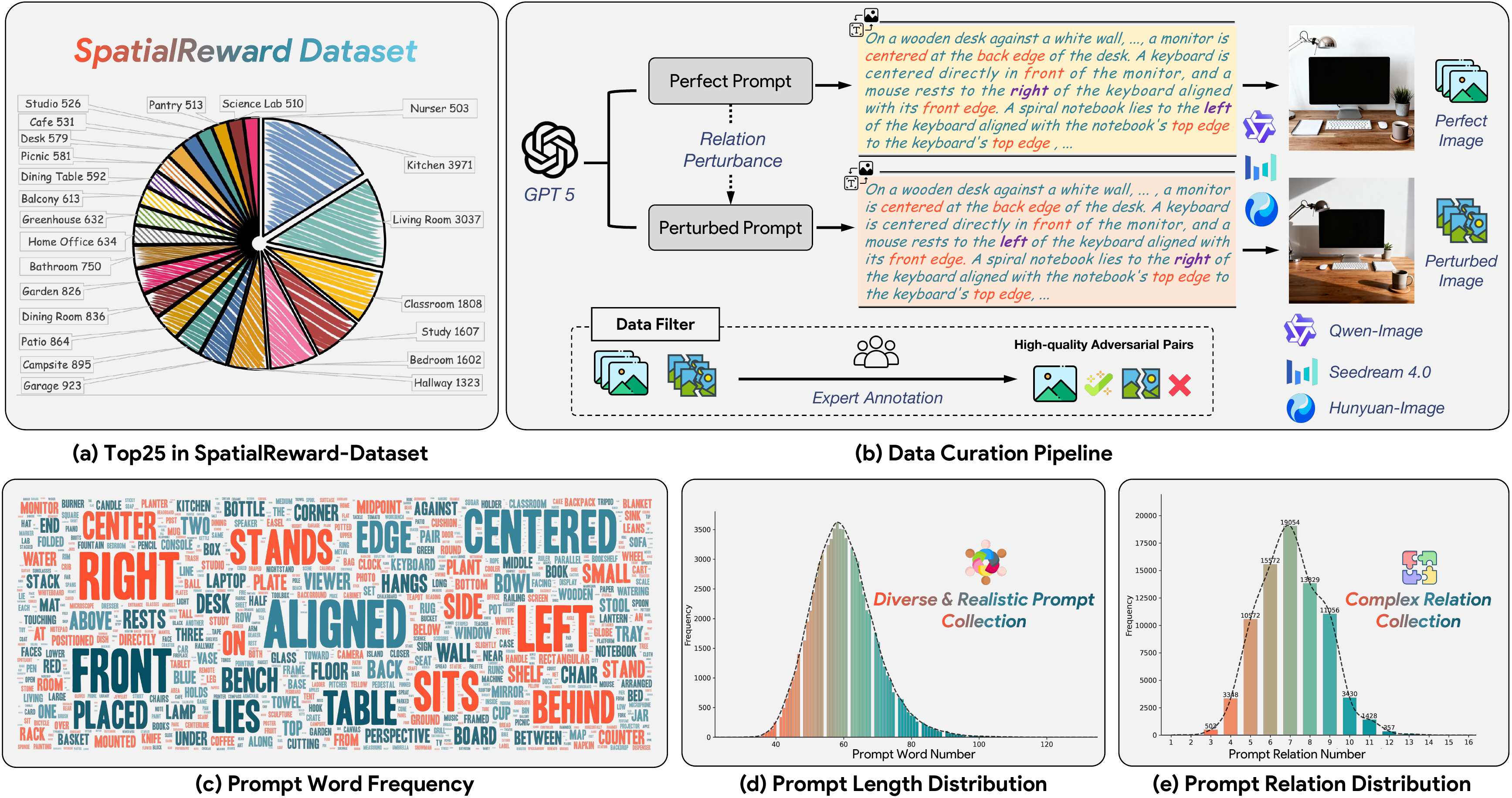}
    \caption{Overview of our SpatialReward-Dataset.}
    \label{fig:spatial_dataset}
\end{figure*}

\section{Related Works}
\label{sec:related_work}

\subsection{Reward Model in T2I models}
\label{sec:related_work:reward_model}
% 先阐述reward model的优势
Reward models are crucial for the success of reinforcement learning (RL) by providing the high-quality signals required for policy optimization.
HPSv2~\cite{wu2023hpsv2}, Pickscore~\cite{kirstain2023pickscore}, Aesthetic score~\cite{burger2023laion}, and the works~\cite{li2025science-t2i,zhang2024learning-hp-t2i, lin2024vqascore} fine-tune CLIP-based model with human preference data, and HPSv3~\cite{ma2025hpsv3} and UnifiedReward~\cite{wang2025unified} employ Vision-Language Models (VLMs) as the backbone for reward generation in the text-to-image task.
However, while these models excel at assessing aesthetic quality and overall text-image alignment, they often lack the fine-grained capacity to comprehend and evaluate complex spatial relationships across multiple objects. 
This limitation can lead to generations that are semantically plausible but compositionally incorrect, which motivates our development of a specialized reward model that focuses on spatial understanding.
% ,z,kirstain2023pickscore,burger2023laion

% \begin{figure}[t!]
%     \centering
%     \includegraphics[width=0.9\linewidth]{Figures/fig_data_relation_sum.pdf}
%     \caption{A placeholder figure using example-image-duck.}
%     \label{fig:relation_sum}
% \end{figure}

\subsection{Reinforcement Learning in Image Generation}
\label{sec:related_work:rl}
% 先阐述rl在image gen的发展，然后从ppo和dpo，再到deepseek-r1的grpo的insipre，再提出与flowmatching结合
% 最后聚焦于，我们的方法提出了一个具有空间感知和指令跟随能力的reward model
Reinforcement learning has been effectively applied in diffusion models to improve generation quality.
Proximal Policy Optimization (PPO)~\cite{schulman2017proximal-policy-opt} and Direct Perference Optimization (DPO)~\cite{rafailov2023direct-dpo} originally developed for large language models, have been successfully adapted to diffusion-based generation~\cite{jiang2025t2i,li2025science-t2i,zhu2025dspo,ren2024-dppo,xu2023imagereward}, improving task alignment and controllability.
Building upon this, to pursue a more stable and efficient optimization process,  Flow-GRPO~\cite{liu2025flow-grpo}, Dance-GRPO~\cite{xue2025dancegrpo} and others~\cite{wang2025pref-grpo,li2025mixgrpo,wang2025grpoguard,he2025tempflow-grpo,zhou2025-g2rpo} have integrated the flow model with Group Relative Policy Optimization (GRPO)~\cite{guo2025deepseek-r1}.
They transform deterministic ordinary differential equation (ODE) sampling into stochastic differential equation (SDE)~\cite{song2020score-sde,albergo2023stochastic-sde-flow} to facilitate policy exploration.
Our work differs by introducing a spatially-aware reward model tailored for spatial understanding in image generation, providing reliable feedbacks.

\newcommand{\algcomment}[1]{%
    \Statex \hspace{-\algorithmicindent} \textcolor{gray}{\textit{#1}}%
}

% ================= BEGIN DOCUMENT =================
\section{Dataset}
\label{sec:dataset}

Building on  VideoAlign~\cite{liu2025improving}, which demonstrates that preference learning outperforms pointwise score regression for reward training, we introduce the carefully curated adversarial \textsc{SpatialReward-Dataset} as the foundation for subsequent reward training.
% designed to evaluate spatial understanding in image generation—specifically, the accuracy of spatial relationships described in the prompts.

As illustrated in Figure~\ref{fig:spatial_dataset}, we construct the \textsc{SpatialReward Dataset} comprising 80K adversarial pairs. In Figure~\ref{fig:spatial_dataset}(a), we illustrate the diverse real-world scenarios used in our data construction.  To minimize the influence of other attributes, such as aesthetic differences across image generation models, we generate each preference pair using a single image generation model with distinct prompts.
Specifically, we first use GPT-5 to create a set of initial prompts featuring complex spatial relationships among multiple objects. We then employ GPT-5 to perturb these clean prompts by modifying one or more spatial relations (e.g., moving an object from left to right, swapping relative positions of objects) while keeping the remaining spatial relationships unchanged. Under this setup, the images generated from the original, unperturbed prompts serve as the \textit{perfect} images, whereas those generated from the perturbed prompts act as the \textit{perturbed} images.

For data construction, we employ sevaral state-of-the-art image generation models Qwen-Image~\cite{wu2025qwen-image} and HunyuanImage-2.1~\cite{HunyuanImage-2.1} and seedream4.0~\cite{gao2025seedream}, which demonstrate strong text–image alignment capabilities, thereby reducing extensive data filtering. Each data pair is manually reviewed and validated by human annotators to filter out cases where the \textit{perfect} image fails to fully align with the complex spatial constraints described in the prompt, ensuring high data quality for reward training. More details of the construction of the SpatialReward-Dataset are provided in the \textit{Appendix}. As shown in Figure~\ref{fig:spatial_dataset}(d), our carefully curated dataset contains significantly longer prompts compared to GenEval~\cite{ghosh2023geneval}, indicating higher scene complexity. Moreover, in Figure~\ref{fig:spatial_dataset}(e), these prompts involve multiple spatial relationships among objects, leading to a greater degree of spatial complexity and compositional diversity than the simple and template-based constructions presented in the Geneval benchmark.

% \subsection{Preliminary}
% \noindent \textbf{Flow Matching.}
% Flow Matching~\citep{lipman2022flow-matching} learns a vector field to transport samples from a simple prior distribution $X_1$ to a complex data distribution $X_0$. 
% We utilize Rectified Flow~\citep{liu2022flow-matching-rectified-flow}, which defines a simple linear interpolation between a data sample $x_0 \sim X_0$ and a prior sample $x_1 \sim X_1$:
% \begin{equation}
% \label{eq:flow_forward}
% x_t = (1-t)x_0 + tx_1, \quad t \in [0,1].
% \end{equation}
% This path is characterized by a constant target velocity field $v = x_1 - x_0$.
% A neural network $v_\theta(x_t, t, c)$, conditioned on $c$ , is trained to approximate this target field by minimizing the flow matching objective:
% \begin{equation}
% \label{eq:flow_matching}
% \mathcal{L}_{FM}(\theta) = \mathbb{E}_{t,x_0 \sim X_0,x_1 \sim X_1}[||v - v_{\theta}(x_t, t, c)||_2^2].
% \end{equation}
% Inference is performed by solving a deterministic ODE for the forward process:
% \begin{equation}
% \label{eq:forward}
%     dx_t = v_{\theta}(x_t, t, c)dt.
% \end{equation}

\section{Method: SpatialScore}
\label{sec:method}
% \subsection{Dataset: SpatialReward-Dataset}
% \label{sec:method:dataset}
\subsection{Architecture}
Recently, Visual Language Models (VLMs) have achieved remarkable progress and been widely applied to various vision–language tasks such as grounding~\cite{peng2023kosmos, zhang2024llava, munasinghe2025videoglamm} and segmentation~\cite{lai2024lisa, yang2023lisa++, ren2024pixellm}. Their success largely stems from training on massive web-scale datasets and learning highly generalizable representations.

Building upon the strong representational power of VLMs as feature extractors, we adopt a VLM as the backbone of our reward model. Specifically, we use Qwen2.5-VL-7B~\cite{bai2025qwen2.5vl} as the backbone $H_{\phi}$ to extract features from both images and text, while replacing the original language modeling head with a new linear reward head $R_{\phi}$ that projects the features to predict the reward value. Our reward model is trained on preference pairs consisting of a preferred image $y_{w}$ as the “winner” and a less preferred image $y_{l}$ as the “loser”, given an instruction prompt $c$. The reward score $s$ for a generated image is computed as:
 \begin{equation}
s \;=\; R_{\phi}\!\left(H_{\phi}(c,\, y)\right).
\label{arch}
\end{equation}

For each preference pair, the scores $s_{w}$ and $s_{l}$ are obtained by feeding the preferred image $y_{w}$ and the less preferred image $y_{l}$ into the model, respectively.

\subsection{Reward training}
To build our \textsc{SpatialScore} reward model, we fine-tune Qwen2.5-VL-7B using the LoRA~\cite{hu2022lora} to preserve the inherent knowledge priors of the model.
In our \textsc{SpatialReward-Dataset} setup, each training example is represented as a triplet $(c, y_{w}, y_{l})$, where $y_{w}$ and $y_{l}$ correspond to the perfect image and the perturbed
image generated from the perturbed prompt, respectively.

Inspired by HPSv3~\cite{ma2025hpsv3}, which models the reward score as a probability distribution for more robust ranking instead of directly outputting a deterministic value, we adopt a Gaussian distribution $s \sim \mathcal{N}(\mu, \sigma^{2})$ to model the final reward score, where $\mu$ and $\sigma$ denote the mean and standard deviation, respectively.
Specifically, within the VLM instruction setup, we insert a special token \texttt{<reward>} at the end of the full prompt, allowing it to attend to both the image and text representations. The final-layer embedding of this special token is then mapped to $\mu$ and $\sigma$ through the reward head $R_{\phi}$, implemented as a multilayer perceptron (MLP). In this way, we model the output score by sampling from this one-dimensional Gaussian distribution.
% 这里的Gaussian需要大写么？

Finally, the training process involves two independent forward passes of the reward model for each triplet sample $(c, y_{w}, y_{l})$ to obtain the reward scores of $y_{w}$ and $y_{l}$. 
The reward model is optimized following the Bradley-Terry model~\cite{bradley1952rank-Bradley-Terry} by minimizing the negative log-likelihood of the ground-truth preference using a binary cross-entropy loss:
\begin{equation}
    P(y_{w} \succ y_{l} \mid c) = 
    \sigma\!\Big(R_{\phi}\!\left(H_{\phi}(y_{w}, c)\right) - 
    R_{\phi}\!\left(H_{\phi}(y_{l} , c)\right)\Big),
\end{equation}
\begin{equation}
    \mathcal{L}_{\text{Reward}}(\theta) = 
    \mathbb{E}_{c, y_{w}, y_{l}}\!\big[-\log P(y_{w} \succ y_{l} \mid c)\big].
\end{equation}

Here, $\sigma$ denotes the sigmoid function, which constrains the output to a probability value within the interval $[0, 1]$, and $\theta$ represents the trainable parameters of the reward model.
Through the optimization of $\mathcal{L}_{\text{Reward}}$, our reward model learns to assign higher scores to the preferred images over less-preferred ones.
Once trained, SpatialScore serves as the reward model for online reinforcement learning, further enhancing spatial understanding in image generation.

\section{SpatialScore in Image Generation}
\label{sec:method:spatial_reward}
With the high-fidelity and specialized reward model \textsc{SpatialScore}, we then leverage it as a direct reward signal for online reinforcement learning  fine-tuning in order to validate the effectiveness of our reward model.
% 搜了一下preceed to好像不能表示情态、而是一个动作的执行，感觉文章还是得用then？

\begin{figure*}[!th]
    \centering
    \includegraphics[width=\linewidth]{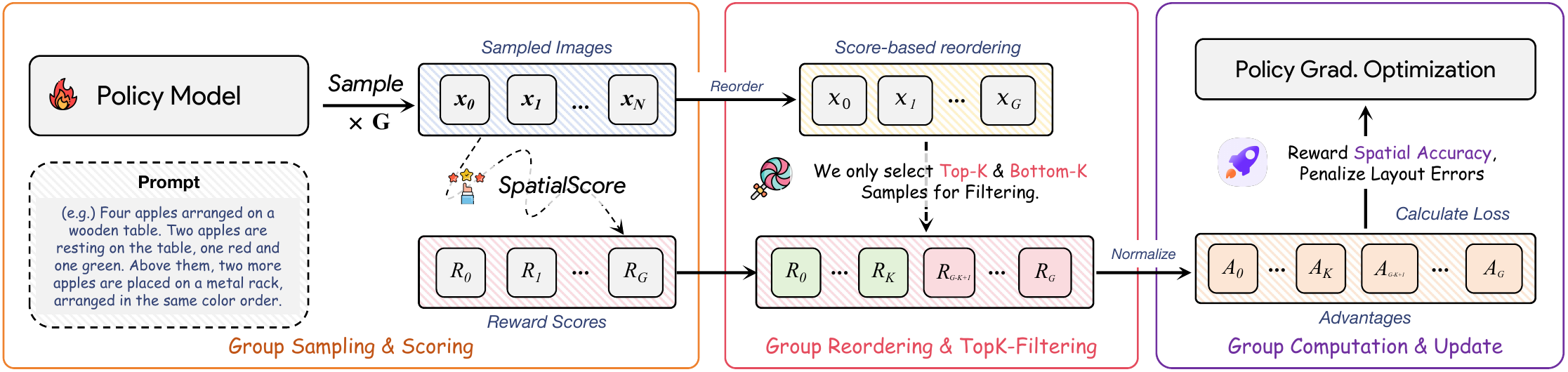}
    \caption{\textbf{GRPO training pipeline for enhancing spatial unserstanding.} 
   We first samples a group of images from the policy model and uses our specialized SpatialScore to rate their spatial accuracy. 
    After ranking based on these scores, we  select the top-$k$ most accurate and bottom-$k$ least accurate examples and convert these scores into advantage signals. 
    The policy model is updated via policy gradient optimization to directly reward correct spatial layouts and penalize errors, thereby enhancing the base model’s spatial understanding.}
    \label{fig:grpo_pipeline}
\end{figure*}

We choose FLUX.1-dev~\cite{flux2024} as our base model for image generation due to its advanced performance and strong support for handling long text inputs, which aligns well with the complex prompt setting of our \textsc{SpatialReward- Dataset}. Moreover, FLUX.1-dev has not undergone the post-training stage, making it an ideal choice to fairly evaluate the  potential gains introduced by our reward model.
As shown in the Figure~\ref{fig:grpo_pipeline}, we employ the GRPO algorithm from FlowGRPO~\cite{liu2025flow-grpo} and leverage our reward model to provide reliable feedback for optimizing the base generation model for spatial understanding.

\begin{figure}[t!]
    \centering
    \includegraphics[width=\linewidth]{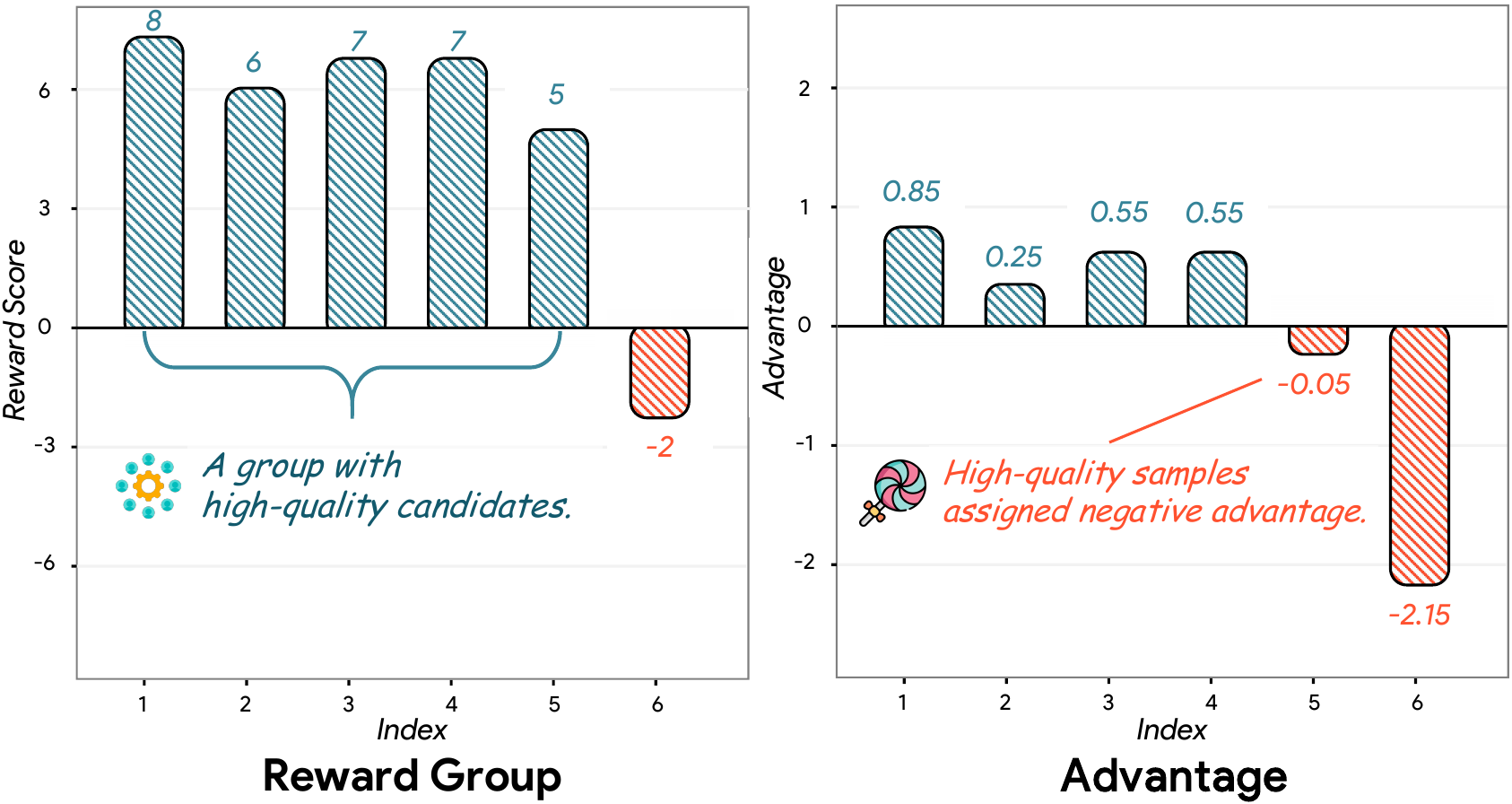}
    \vspace{-15pt}
    \caption{\textbf{Advantage bias.} For easy prompts with many high-reward samples, some high-quality samples often obtain negative advantages due to the high group mean.}
    \vspace{-15pt}
    \label{fig:bias}
\end{figure}

Specifically, as an online RL algorithm, GRPO~\cite{shao2024deepseekmath} requires diverse samples within a group to optimize the policy and evaluate the reward uplift direction.
However, flow matching inherently employs a deterministic ODE for sampling, whereas RL relies on stochasticity for policy exploration.
Following Flow-GRPO~\cite{liu2025flow-grpo}, we convert the deterministic ODE into an equivalent SDE that shares the same marginal distribution.
The resulting SDE can be discretized using the Euler–Maruyama scheme as follows:
\begin{equation}
\begin{aligned}
x_{t+\Delta t}
&= x_t +
\left[
v_\theta(x_t, t) +
\frac{\sigma_t^2}{2t}
\left(
x_t + (1 - t)v_\theta(x_t, t)
\right)
\right]\Delta t \\
&\quad + \sigma_t \sqrt{\Delta t}\,\epsilon
\end{aligned}
\end{equation}

where $\epsilon \sim \mathcal{N}(0, I)$ denotes standard Gaussian noise, $\sigma_t$ represents the injected noise level, and $v_{\theta}(x_t, t)$ denotes the estimated velocity field.

For each  prompt $c$ sampled from our \textsc{SpatialReward-Dataset}, 
the base model as the policy $\pi_{\theta}$ generates a group of $G$ images 
$\{x^{i}_{0}\}_{i=1}^{G}$ through SDE sampling. 
Our reward model \textsc{SpatialScore} then evaluates each generated image 
$x^{i}_{0}$ conditioned on $c$ and assigns a reward score $R(x^{i}_{0}, c)$. 
Within each group, the advantage $A^{i}$ of the $i$-th image is computed 
by normalizing its reward with respect to the group mean and standard deviation:
\begin{equation}
A^i
= \frac{ R(x_i^{0},\, c)
- \operatorname{mean}\!\big(\{\, R(x_0^{i},\, c) \,\}_{i=1}^{G}\big) }
{ \operatorname{std}\!\big(\{\, R(x_0^{i},\, c) \,\}_{i=1}^{G}\big)} \, .
\end{equation}

Based on the preliminary findings in {Flow-GRPO~\cite{liu2025flow-grpo}, a sufficiently large group size is essential to ensure diverse sampling and stabilize training.
However, in our practical experiments, as online RL training progresses, prompts of varying difficulty may cause biased advantage estimation. Specifically, easy prompts tend to accumulate a large number of successful samples with high reward scores within a group, while difficult prompts often produce samples with generally low rewards.
As illustrated in Figure~\ref{fig:bias}, for an easy prompt, the group-wise normalization may yield a high mean value, which in turn assigns negative advantages to some high-quality samples, producing optimization gradients that deviate from the intended reward-improvement direction. Conversely, hard prompts also yield advantage bias.

To mitigate advantage biases across prompts of varying difficulty, 
we propose a top-$k$ filtering strategy. 
Specifically, for a sampled group of $G$ images $\{x^{i}_{0}\}_{i=1}^{G}$ generated by the policy $\pi_{\theta}$, 
we rank them based on the reward scores estimated by our reward model, 
obtaining a sorted sequence $\{\hat{x}^{i}_{0}\}_{i=1}^{G}$. 
To balance the reward distribution within the group and further mitigate advantage bias, 
we select both the top-$k$ and bottom-$k$ samples to form a index subset, i.e., $S=\{1,2,\ldots,k,\, G-k+1,\ldots,G\}$,
which is then used to compute the group mean and standard deviation. 
During policy updates, only these selected samples in this subset $S$ are utilized for training.

\begin{table*}[t]
  \centering
  \caption{Pairwise-accuracy comparisons on the reward evaluation benchmark. “1 Pert.” and “2–3 Pert.” denote subsets with one or two–three spatial perturbations applied to the perfect prompts when constructing the perturbed prompts.}
  \vspace{-0.2cm}
  \renewcommand{\arraystretch}{1.35}
  \setlength{\tabcolsep}{6pt}
  \resizebox{\textwidth}{!}{%
  \begin{tabular}{l*{12}{c}}
    \toprule
    \multirow{3}{*}{\textbf{Setting}} &
    \multicolumn{6}{c}{\textbf{Image Reward Models}} &
    \multicolumn{3}{c}{\textbf{Qwen2.5-VL Series}} &
    \multicolumn{2}{c}{\textbf{Proprietary Models}} & 
    \multirow{3}{*}{\textbf{SpatiaScore}}\\
    \cmidrule(lr){2-7} \cmidrule(lr){8-10} \cmidrule(l){11-12}
    & \makecell{Image\\Reward} & Pickscore & HPSv2.1 & VQAScore
    & \makecell{Unified\\Reward} & HPSv3
    & \makecell{7B} & \makecell{32B} & \makecell{72B}
    & GPT-5 & \makecell{Gemini-\\2.5 pro}   \\
    \midrule
    1 Pert. & 0.439 & 0.461 & 0.433& 0.567 & 0.583 & 0.606 & 0.572 & 0.644 & 0.711 & 0.855 & 0.933 & \textbf{0.939} \\
    2--3 Pert. & 0.513 & 0.551 & 0.491 & 0.638 & 0.627 & 0.697 & 0.632 & 0.724 & 0.816 & 0.924 & 0.968 & \textbf{0.978} \\
    \midrule
    \rowcolor{gray!10} 
    Overall
      & 0.479 & 0.509 & 0.463 & 0.603
      & 0.605 & 0.652
      & 0.602 & 0.685 & 0.764
      & 0.890 & 0.951 & \textbf{0.958} \\
    \bottomrule
  \end{tabular}}
  \label{tab:peference_accuracy}
\end{table*}

\begin{table*}[t]
  \centering
  \caption{Detailed comparisons on SpatialScore, DPG-Bench, TIIF-Bench (short/long), and UnigenBench++ (short/long). * denotes training with Geneval as the reward model.
BR, AR, and RR denote basic relation, attribute+relation, and relation+reasoning.
Lay-2D/3D refer to layout-2D/3D. Unibench denotes UniGenBench++.}
  \vspace{-0.2cm}
  \renewcommand{\arraystretch}{1.36}
  \setlength{\tabcolsep}{6pt}
  \resizebox{\textwidth}{!}{%
  \begin{tabular}{l|*{12}{c}}
    \toprule
    \multirow{2}{*}{\textbf{Method}} &
    \multirow{2}{*}{\textbf{SpatialScore}} &
    \multicolumn{1}{c}{\textbf{DPG-bench}} &
    \multicolumn{3}{c}{\textbf{TIIF-bench-short}} &
    \multicolumn{3}{c}{\textbf{TIIF-bench-long}} &
    \multicolumn{2}{c}{\textbf{Unibench(short)}} &
    \multicolumn{2}{c}{\textbf{Unibench(long)}} \\
    \cmidrule(lr){3-3}\cmidrule(lr){4-6}\cmidrule(lr){7-9}\cmidrule(lr){10-11}\cmidrule(l){12-13}
    & & Relation-Spatial
      & BR & AR & RR
      & BR & AR & RR
      & Lay-2D & Lay-3D
      & Lay-2D & Lay-3D \\
    \midrule
    
  %   SD3.5-M
  %     & --   & --     
  %     & --   & --   & -- 
  %     & --   & --   & --
  %     & --   & --
  %     & --   & -- \\
  %     SD3.5-M
  %     & --   & --     
  %     & --   & --   & -- 
  %     & --   & --   & --
  %     & --   & --
  %     & --   & -- \\
  % \midrule
    Flux.1-dev~\cite{flux2024}
      & 2.18
      & 0.871
      & 0.769   & 0.608  & 0.584
      & 0.758 & 0.677 & 0.645
      & 0.766 & 0.667
      & 0.819  & 0.742 \\
    Flow-GRPO*~\cite{liu2025flow-grpo}
      & 3.01
      & 0.742
      & 0.851 & 0.652  & 0.621
      & 0.577 & 0.510 & 0.482
      & 0.726 & 0.635 
      & 0.445 & 0.405   \\
    \rowcolor{gray!10} 
    Ours
      & \textbf{7.81}
      & \textbf{0.932}
      & \textbf{0.875} & \textbf{0.700} & \textbf{0.647}
      & \textbf{0.845} & \textbf{0.715} & \textbf{0.675}
      & \textbf{0.875} & \textbf{0.773}
      & \textbf{0.891} &\textbf{0.801} \\
    \bottomrule
  \end{tabular}}
  \label{tab:bench_grouped}
\vspace{-0.2cm}
  
\end{table*}

The GRPO algorithm then optimizes the policy by
minimizing the following objective:
\begin{align}
\mathcal{L}_{\text{GRPO}}(\theta)
= & \frac{1}{|S|}\sum_{i\in S} \frac{1}{T} \sum_{t=0}^{T-1}
\min\left( r_t^i(\theta)\,A_t^i,\right. \notag\\
&\left. \text{clip}\big(r_t^i(\theta), 1-\epsilon, 1+\epsilon\big)\,A_t^i \right).
\label{eq:policy}
\end{align}

where $
r_t^i(\theta)=\frac{
    p_\theta(\hat{x}_{t-1}^i \mid \hat{x}_t^i, c)
}{
    p_{\theta_{\text{old}}}(\hat{x}_{t-1}^i \mid \hat{x}_t^i, c)
}$. An additional KL-divergence penalty term $D_{\mathrm{KL}}\!\left(\pi_{\theta}\,\|\,\pi_{\mathrm{ref}}\right)$ is introduced to regularize the policy $\pi_{\theta}$ and prevent excessive deviation from the  
reference policy $\pi_{\text{ref}}$. This ensures that the generation model is directly optimized toward improved spatial understanding, driven by feedback from our reward model.

Our GRPO optimization reduces the number of function evaluations (NFE) required for updating the policy $\pi_{\theta}$ compared to optimizing over all samples within each group. Following the empirical results of Flow-GRPO~\cite{liu2025flow-grpo}, which indicate that directly reducing the sampling group size might lead to training collapse for GRPO, we maintain a consistent group size with Flow-GRPO to ensure sufficient sample diversity during sampling.

\section{Experiments}
\label{sec:exps}

\subsection{Experimental Settings}
\noindent \textbf{Implementation Details.}
For reward model training, our \textsc{SpatialScore} is built by fine-tuning Qwen2.5-VL-7B~\cite{bai2025qwen2.5vl} with LoRA~\cite{hu2022lora} on our curated \textsc{SpatialReward-Dataset}, which contains 80K preference pairs.
The training completes within one day on eight NVIDIA H20 GPUs, using a learning rate of $2\times10^{-6}$ and a gradient accumulation batch size of 32.
For RL training, we apply the trained \textsc{SpatialScore} as the reward model to fine-tune the base model Flux.1-dev~\cite{flux2024} which supports long-text prompts for complex scenes in an online RL setup.
Following Flow-GRPO~\cite{liu2025flow-grpo}, we adopt LoRA-based fine-tuning for GRPO training, with a LoRA rank of 32, a learning rate of $3\times10^{-4}$, an importance clipping range of $1\times10^{-4}$, a group size of 24, and a KL-penalty coefficient of 0.01.
The online RL training is conducted on 32 NVIDIA H20 GPUs. 
% Additional training details are provided in the \textit{Appendix}.

\begin{figure*}[t!]
    \centering
    \includegraphics[width=0.95\linewidth]{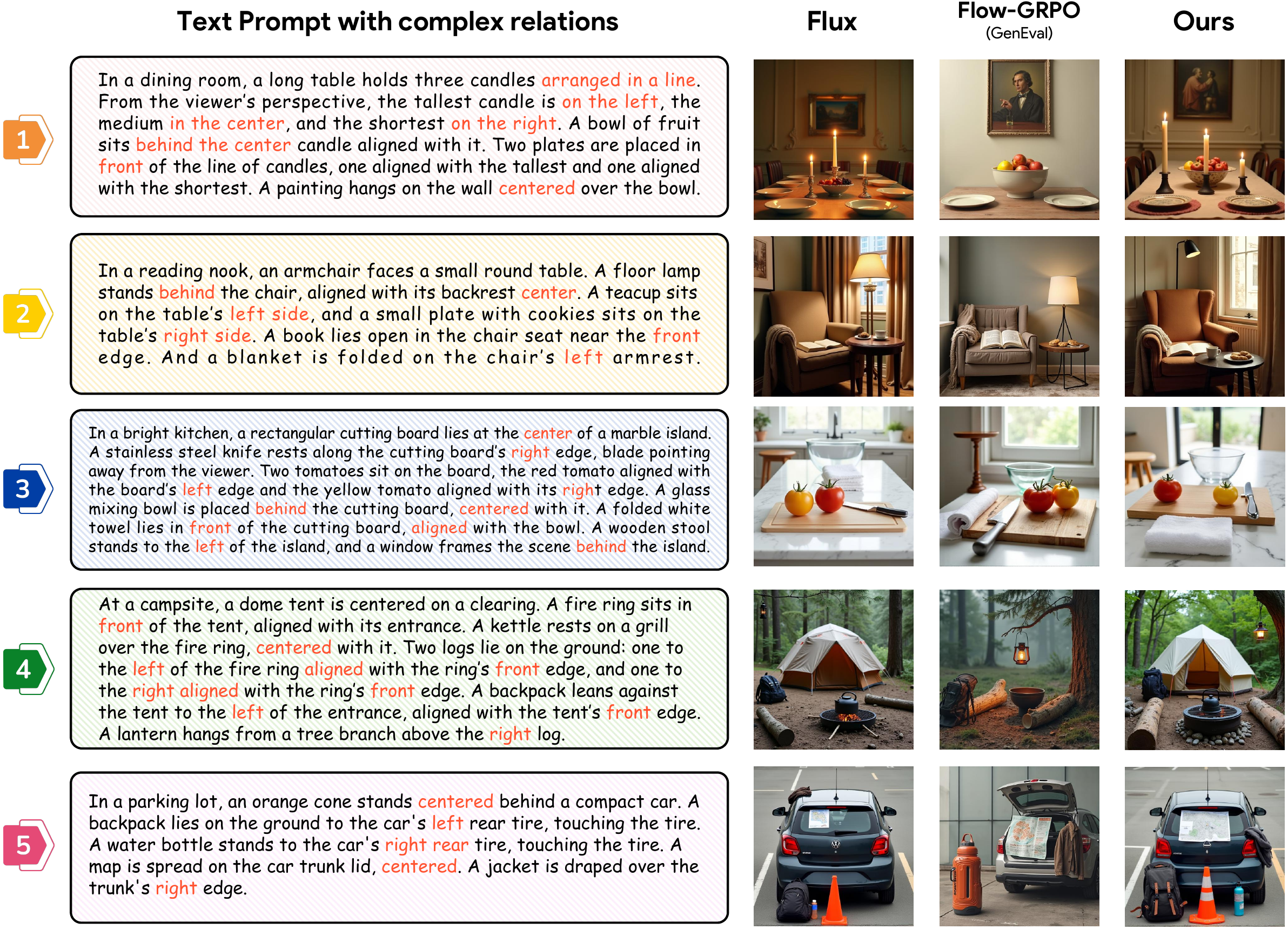}
    \vspace{-5pt}
    \caption{
    Qualitative comparison on prompts with complex spatial relationships across multiple objects. 
    }
    \vspace{-0.4cm}
    \label{fig:visualize}
\end{figure*}

\noindent \textbf{Evaluation Benchmarks for SpatialScore.}
For reward model evaluation, we construct a high-quality and diverse benchmark consisting of 365 preference pairs.
Following a process similar to \textsc{SpatialReward-Dataset}, we generate \textit{perturbed}
prompts by perturbing the original \textit{perfect} prompts.
Each prompt pair is then generated to images using Qwen-Image~\cite{wu2025qwen-image},  HunyuanImage-2.1~\cite{HunyuanImage-2.1}  and Seedream-4.0~\cite{gao2025seedream}, followed by meticulous human review and verification to ensure annotation reliability.
We evaluate a wide range of leading models on this benchmark using overall preference accuracy as the metric.
The evaluation includes proprietary models such as GPT-5~\cite{openai2025gpt5} and Gemini-2.5~\cite{comanici2025gemini}, advanced open-source VLMs from the Qwen2.5-VL~\cite{bai2025qwen2.5vl} series from 7B to 72B, and several existing image reward models, including PickScore~\cite{kirstain2023pickscore}, ImageReward~\cite{xu2023imagereward}, UnifiedReward~\cite{wang2025unified}, and the HPS series~\cite{wu2023hpsv2,ma2025hpsv3}. For UnifiedReward, we employ the Qwen2.5-VL-based models with superior performance.

\noindent \textbf{Evaluation Benchmarks in image generation.}
For evaluating the spatial understanding of image generation models, we first employ our proposed reward model \textsc{SpatialScore} to assess in-domain performance in spatial reasoning.
Beyond in-domain evaluation, we further adopt several out-of-domain benchmarks designed to measure text–image alignment, from which we select the spatial-aware sub-dimensions to specifically evaluate spatial understanding in image generation.
In particular, we utilize DPG-Bench~\cite{hu2024dpg}, which focuses on complex text-to-image alignment; TIIF-Bench~\cite{wei2025tiif}, an extension of T2I-Compbench++~\cite{wang2025unigenbench++} to long prompts, evaluated by GPT-4o~\cite{GPT}; and the recently released UniGenBench++~\cite{wang2025unigenbench++}, which is assessed by Gemini-2.5 Pro~\cite{comanici2025gemini}.
% and provides reliable multi-dimensional alignment evaluation.

\subsection{Reward Model Performance}
\label{sec:exp:quan_exps}
Building upon our carefully curated \textsc{SpatialReward-Dataset}, we propose a state-of-the-art specialized reward model \textsc{SpatialScore} for evaluating spatial understanding: the accuracy of complex spatial relationships among multiple objects. 
As shown in Table~\ref{tab:peference_accuracy}, we compare the accuracy of preference prediction against a broad set of baselines, including human-preference reward models, reward models tailored for text–image alignment, advanced open-source vision language models (VLMs), and the proprietary models. To analyze robustness across difficulty levels, we split our built benchmark by the number of spatial perturbations applied to the  \textit{perfect} prompt to construct the \textit{perturbed} prompt: a 1-perturbation subset and a 2-3 perturbation subset, while keeping all other spatial relations unchanged during data construction.
% 这个修改主要是原来的 —specifically, 真的比较奇怪，然后直接用冒号表达解释

As shown in Table~\ref{tab:peference_accuracy}, proprietary models achieve consistently higher preference-prediction accuracy than both open-source vision language models (VLMs) and existing image-reward models on our benchmark. Leading proprietary models such as GPT-5~\cite{openai2025gpt5} and Gemini-2.5  Pro~\cite{comanici2025gemini} constitute the top tier, attaining the accuracies from 0.89 to 0.95, which shows their strong zero-shot spatial understanding for image generation.
However, their high per-query cost makes them impractical for the frequent evaluations required by online RL.
By contrast, existing image-reward models, which consider text-image alignment, exhibit limited ability to assess multi-object spatial reasoning, attaining limited pairwise accuracies. Open-source VLMs also exhibit clear limitations. Although the Qwen2.5-VL series shows a scaling trend in spatial understanding from 7B to 72B parameters, the 72B variant reaches only 0.76 pairwise accuracy, well below proprietary models, and the 7B and 32B models perform even worse. These results indicate that current advanced open-source VLMs are not yet reliable providers of rewards for complex spatial reasoning.

In contrast, our \textsc{SpatialScore} with 7B parameters achieves state-of-the-art results on the reward benchmark, reaching a pairwise accuracy of 95.77\%. It surpasses strong proprietary models, including GPT-5 and Gemini-2.5 Pro, on spatial understanding across multiple objects. These results indicate that our specialized reward model can provide a reliable and stable reward signal for online RL.

\subsection{Applying SpatialScore for Online RL}
\label{sec:exp:qua_exps}
\textbf{Quantitative results.} Guided by our specialized reward model \textsc{SpatialScore}, we perform online RL training using Flux.1-dev as the base model. As shown in Table~\ref{tab:bench_grouped}, we compare with the original base model and a variant from Flow-GRPO trained on GenEval for compositional image generation across multiple benchmarks. On the in-domain \textsc{SpatialScore} evaluation, our approach improves the score from 2.18 to 7.81, demonstrating the effectiveness of reward-guided training. To further assess spatial understanding, we evaluate several text–image alignment benchmarks and select their spatial-aware subdimensions. Our RL training yields consistent gains on both short- and long-prompt settings. In contrast, the model trained with Flow-GRPO on the rule-based GenEval shows some improvement on short prompts but degrades markedly on long-prompt settings, indicating limited generalization to long prompts with complex multi-object spatial relationships.

\begin{figure}[t!]
    \centering
    \includegraphics[width=0.95\linewidth]{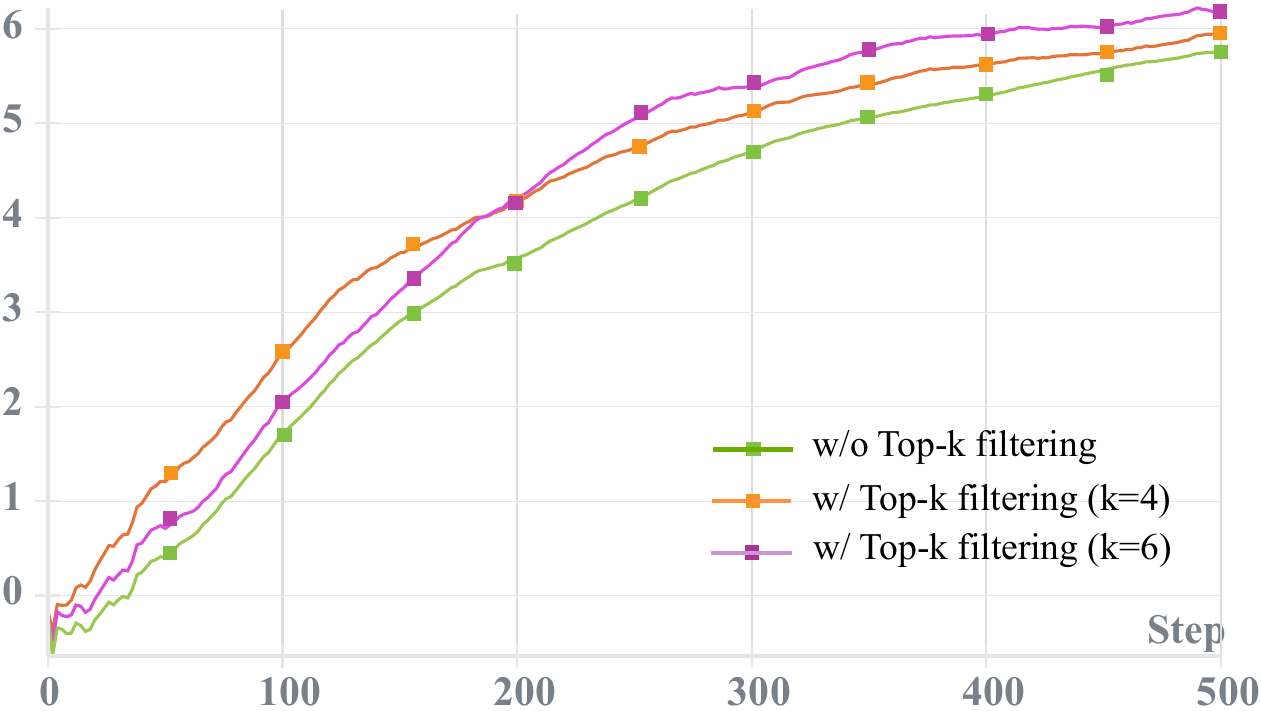}
    \vspace{-5pt}
    \caption{Reward training curves for ablations on top-$k$ filtering}
    \label{fig:log_wandb}
    \vspace{-5pt}
\end{figure}

\begin{table}[!t] % 双栏跨全幅可改为 table* 并把 \columnwidth 换成 \textwidth
  \centering
  \vspace{-5pt}
  \caption{Full-dimensional comparison on DPG-Bench. The best Flux variants are in \textbf{bold}, and * denotes training with GenEval.}
  \vspace{-5pt}
  \renewcommand{\arraystretch}{1.5}
  \setlength{\tabcolsep}{8pt}
  \resizebox{\columnwidth}{!}{%
  \begin{tabular}{l|ccccc|c}
    \toprule
    \textbf{Method} & \textbf{Global} & \textbf{Entity} & \textbf{Attribute} & \textbf{Relation} & \textbf{Other} & \textbf{Overall} \\
    \midrule
    GPT-Image-1 & 88.89 & 88.94 & 89.84 &  92.63 &  90.96 & 85.15 \\
    \midrule
    
    Flux.1-dev         & 84.70 & 87.29 & 89.29 & 89.44 & 89.46 & 82.91 \\
    Flow-GRPO* & 69.20 & 72.37 & 69.23 & 73.26 & 69.72 & 57.02 \\
    \rowcolor{gray!10} 
    Ours               & \textbf{87.65} & \textbf{90.56} & \textbf{91.20} & \textbf{91.58} &\textbf{ 86.43} & \textbf{85.03} \\
    \bottomrule
  \end{tabular}}
  \label{tab:dpg}
  \vspace{-15pt}
\end{table}
% \vspace{-5pt}

\noindent \textbf{Qualitative results.} As shown in Figure~\ref{fig:visualize}, we present qualitative comparisons against the original Flux.1-dev model and the variant trained with Flow-GRPO on the rule-based GenEval benchmark. 
% Across a diverse set of prompts involving multiple spatial relationships, 
Our RL-trained model using the specialized \textsc{SpatialScore} exhibits improved spatial understanding, producing images that more faithfully reflect the complex spatial relationships across multiple objects described in the  prompts. 
In contrast, the Flux.1-dev variant trained with Flow-GRPO on GenEval demonstrates limited ability to generalize to complex spatial compositions and even loses part of the base model’s long-prompt following capability. As shown in Figure~\ref{fig:visualize}, this includes missing key objects such as the candles in example (1) and the tent in example (4). Moreover, due to the reliance on rule-based rewards from object detectors, the GenEval-guided model frequently generates visually implausible artifacts, such as floating maps or jackets, as seen in example (5).

\noindent \textbf{Out-of-domain evaluation.} We further conduct a full-spectrum evaluation on DPG-Bench~\cite{hu2024dpg}, a benchmark designed to assess text–image alignment. As shown in Table~\ref{tab:dpg}, our method yields consistent and substantial improvements over the original Flux.1-dev model across all five major dimensions of DPG-Bench, with gains observed beyond the spatial subdimension. Notably, the overall performance of our RL-enhanced model approaches that of the proprietary GPT-Image-1. In contrast, the Flux  variant trained with Flow-GRPO on GenEval exhibits clear degradation.

% \subsection{User Study}
% \label{sec:exp:user_study}
% There are some sentences to fill this paper. There are some sentences to fill this paper. There are some sentences to fill this paper. There are some sentences to fill this paper. There are some sentences to fill this paper. There are some sentences to fill this paper. There are some sentences to fill this paper. There are some sentences to fill this paper. There are some sentences to fill this paper. There are some sentences to fill this paper.  
% There are some sentences to fill this paper. There are some sentences to fill this paper. There are some sentences to fill this paper. There are some sentences to fill this paper. There are some sentences to fill this paper. There are some sentences to fill this paper. There are some sentences to fill this paper. There are some sentences to fill this paper. There are some sentences to fill this paper. There are some sentences to fill this paper.  

\subsection{Ablation Study}
\label{sec:exp:ablation_study}

\textbf{Ablations on reward model size.} To assess the generalizability of our approach, we train \textsc{SpatialScore} with varying backbone sizes. On our reward evaluation benchmark, \textsc{SpatialScore} improves the accuracy of pairwise preference prediction from 89.1\% to 95.8\%  when scaling the backbone from Qwen2.5-VL-3B to Qwen2.5-VL-7B. More additional results are provided in the Appendix.

\vspace{-5pt}
\begin{table}[!ht]
\centering
\vspace{-5pt}
\caption{Ablations on top-$k$ filtering. NFE per prompt for each training step is reported under a denoising step count of 6.}
\vspace{-0.22cm}
\label{tab:spatial_bench}
\renewcommand{\arraystretch}{1.2}
\resizebox{\linewidth}{!}{
\begin{tabular}{l|cccccc}
\toprule
\multirow{2}{*}{\textbf{Setting}} &
\multirow{2}{*}{\textbf{SpatialScore}} &
 \multicolumn{1}{c}{\textbf{DPG-bench}} &
\multicolumn{2}{c}{\textbf{Unigenbench++ (long)}} &
\multirow{2}{*}{\textbf{NFE}} \\
\cmidrule(lr){3-5}
& &Rel-Spatial & Layout-2D & Layout-3D & \\
\midrule

w/o top-$k$             & 7.73    & 0.919 & \textbf{0.891} & 0.793 & 24*6 \\
\midrule
w/ top-$k$ (k=4)   & 7.71 & 0.916   & 0.882 & 0.796 & \textbf{8*6} \\
w/ top-$k$ (k=6)             & \textbf{7.81}   & \textbf{0.932}   & 0.887& \textbf{0.801} & 12*6 \\
\bottomrule
\end{tabular}
}
\vspace{-5pt}
\end{table}

\vspace{-5pt}

\noindent \textbf{Ablations on top-$k$ filtering.} In online RL training, prompts of varying difficulty can lead to highly imbalanced reward distributions within a  group. 
In particular, easy prompts often produce many high-reward samples, which obtain high group mean and consequently assign negative advantages to some high-quality samples. To mitigate this issue, we apply a top-$k$ filtering strategy that selects the top-$k$ and bottom-$k$ samples within the group to construct a balanced subset and reduce advantage bias. As shown in Figure~\ref{fig:log_wandb}, adding top-$k$ filtering accelerates training compared to the baseline GRPO setup without filtering. When $k=4$, the training exhibits faster early-stage improvement but slows down in later stages due to insufficient sample diversity. In contrast, $k=6$ maintains a better tradeoff between sampling balance and diversity, and we adopt $k=6$ as the default configuration in all experiments. As shown in Table~\ref{tab:spatial_bench}, we observe that using $k=6$ achieves comparable or even superior performance with fewer the number of function evaluations (NFE) in the policy update stage. With a sampling group size of 24 and denoising steps of 6, only 2*6*6 NFEs per prompt when $k=6$ at each step for training stage, whereas the original RL needs 24*6 NFEs.
\section{Conclusion}
\label{sec:conclusion}

In this work, we address the challenge of improving spatial understanding in image generation through online RL. To this end, we first introduce the \textsc{SpatialReward-Dataset}, an 80K-pair preference dataset curated with rigorous human verification. Building on this dataset, we develop \textsc{SpatialScore}, a specialized reward model that provides reliable signals and achieves evaluation accuracy surpassing even proprietary models. Leveraging this high-fidelity reward model and GRPO framework with top-$k$ filtering, we obtain substantial and consistent improvements in spatial reasoning across multiple benchmarks over the base model.

{
    \small
    \bibliographystyle{ieeenat_fullname}
    \bibliography{main}
}

\clearpage
\setcounter{page}{1}
\maketitlesupplementary

% \tableofcontents
% \addtocontents{toc}{\setcounter{tocdepth}{3}}
\addtocontents{toc}{\protect\setcounter{tocdepth}{3}}
\tableofcontents

\section{Experimental Details}

\begin{tcolorbox}[colback=gray!10, 
                    colframe=gray!80,  
                    float*=htbp, 
                    width=\textwidth, 
                    title=\textbf{Instruction Template}, 
                    label={textbox:template},
                    breakable]
% \label{textbox:template}
You are tasked with evaluating a generated image based on Spatial Position Consistency. Please provide a rating from 0 to 10, with 0 being the worst and 10 being the best.

\vspace{+5pt}
\textbf{Spatial Position Consistency:}

Evaluate how well the spatial relationships in the image match the descriptions in the prompt. Focus on the relative positions and arrangements of multiple objects in relation to each other and the background in the image. The following sub-dimensions should be considered:

\begin{itemize}
    \item \textbf{Relative Positioning between objects.} Evaluate if the relative positions between the objects in the image match the description in the prompt. The objects should be placed in the described order and relative positions, without contradictions.
    \item \textbf{Relative Positioning with Background.} Evaluate if the relative positions of all objects with respect to the background in the image match the description in the prompt. 
    \item \textbf{Consistency in Object Attributes.} Check if the properties of objects (such as size and color) at their specified positions align with the descriptions in the prompt. Ensure that the attributes of the objects are consistent with their spatial relationships. 
\vspace{+5pt}  
\end{itemize}
\noindent Textual prompt: \texttt{[text prompt]}

Please provide the overall ratings of this image: \texttt{<|Reward|>}

END
\end{tcolorbox}

\subsection{Reward Model Training}
Our reward model \textsc{SpatialScore} is constructed by fine-tuning Qwen2.5-VL-7B~\cite{bai2025qwen2.5vl} with LoRA~\cite{hu2022lora} on our curated \textsc{SpatialReward-Dataset}, which consists of 80k preference pairs covering a wide range of real-world scenarios. During training, we insert a special token \texttt{<reward>} into the instruction to attend to both visual and textual features. The complete instruction template is shown in the textbox~\ref{textbox:template} on the top of the next page. The final-layer embedding of this special token is projected by the reward head $R_{\phi}$ into the mean $\mu$ and standard deviation $\sigma$ of a Gaussian distribution, from which the final reward score is obtained through sampling. To enhance training stability, we sample 1000 times from the Gaussian distribution of both the preferred image $y_w$ and the perturbed image $y_l$ when computing $\mathcal{L}_{\text{Reward}}$, and use the average of these 1000 score pairs to compute the final loss. Training completes within one day on 8 NVIDIA H20 GPUs, using a learning rate of $2 \times 10^{-6}$, a batch size of 16, and gradient accumulation steps of~2.

\subsection{Reward Model Evaluation}
Similar to the construction of the \textsc{SpatialReward-Dataset}, we build a benchmark comprising 365 preference pairs for reward model evaluation. Each preference pair contains a \textit{prefect} image, generated by a perfect prompt, and a \textit{perturbed} image generated from the corresponding perturbed prompt. Each preference pair undergoes rigorous human review and verification to ensure the reliability and consistency of the annotations. We evaluate a wide range of leading models on this benchmark using overall preference accuracy as the evaluation metric. The evaluation includes proprietary models such as GPT-5~\cite{openai2025gpt5} and Gemini-2.5 Pro~\cite{comanici2025gemini}, advanced open-source VLMs from the Qwen2.5-VL series~\cite{bai2025qwen2.5vl} from 7B to 72B, as well as several existing image reward models: PickScore~\cite{kirstain2023pickscore}, ImageReward~\cite{xu2023imagereward}, UnifiedReward~\cite{wang2025unified}, and the HPS family~\cite{wu2023hpsv2,ma2025hpsv3}. For UnifiedReward, we adopt its Qwen2.5-VL-based variants due to their superior performance. For the evaluation of proprietary models and the Qwen2.5-VL series, we instruct the VLMs to choose between the two images in each preference pair by selecting either 'the first image' or 'the second image'. Considering that model predictions may be sensitive to the presentation order of the images, we mitigate this potential bias by performing the evaluation twice, each time with the image order reversed. We then compute average accuracy across two evaluations to obtain a more robust and reliable performance metric.

\subsection{Applying SpatialScore for Online RL}
To further validate the improved spatial understanding through \textsc{SpatialScore}, we use our reward model for online RL training. Specifically, the reward model is applied to RL fine-tuning the base model, Flux.1-dev~\cite{flux2024}, which supports the long-text prompts necessary to align with our complex spatial scenarios. Following the Flow-GRPO~\cite{liu2025flow-grpo}, we adopt LoRA-based RL fine-tuning and leverage the \textit{perfect} prompts from our curated \textsc{SpatialReward-Dataset} for GRPO training, employing the following hyperparameters: a LoRA rank of 32, a learning rate of $3\times10^{-4}$, an importance clipping range of $1\times10^{-4}$, a sampling group size of 24, and a KL-penalty coefficient of $0.01$. The entire online RL training process is conducted on 32 NVIDIA H20 GPUs.

\subsection{Evaluations for Image Generation}
For evaluating the spatial understanding of image generation models, we first employ our proposed reward model \textsc{SpatialScore} to assess in-domain performance in complex spatial reasoning using the prompts from our reward model evaluation benchmark.
Beyond in-domain evaluation, we further adopt several out-of-domain benchmarks designed to measure text–image alignment, from which we specifically select the spatial-aware sub-dimensions to robustly evaluate spatial understanding in image generation.
In particular, we utilize DPG-Bench~\cite{hu2024dpg}, which focuses on complex text-to-image alignment; TIIF-Bench~\cite{wei2025tiif}, an extension of T2I-Compbench++~\cite{wang2025unigenbench++} to long prompts, evaluated by the proprietary model GPT-4o~\cite{GPT}; and the recently released UniGenBench++~\cite{wang2025unigenbench++}, which is assessed by the powerful leading model Gemini-2.5 Pro~\cite{comanici2025gemini}.
and provides reliable multi-dimensional alignment evaluation.

\section{Dataset Construction}

\subsection{Preference Pair Construction}
For our curated \textsc{SpatialReward-Dataset}, we construct 80k adversarial preference pairs for subsequent reward model training. 
To minimize the influence of other factors, such as aesthetic differences across image generation models, we generate each preference pair using a single image generation model while varying only the prompts. Specifically, we first employ \textsc{GPT-5}~\cite{openai2025gpt5} to create an initial set of prompts featuring complex spatial relationships among multiple objects. We then use GPT-5 to perturb these clean prompts by modifying one or more spatial relations (e.g., moving an object from left to right, swapping the relative positions of objects) while keeping the remaining relationships unchanged. Under this setup, images generated from the original, unperturbed prompts serve as the \textit{perfect} images, whereas those generated from the perturbed prompts act as the \textit{perturbed} images, thereby forming a complete preference pair. As illustrated in Figure~\ref{fig:supp:more_data_pairs1} and Figure~\ref{fig:supp:more_data_pairs2}, we show several representative examples of preference pairs to visualize our curated SpatialReward-Dataset.

To further enhance the diversity of our dataset, we employ several state-of-the-art image generation models Qwen-Image\cite{wu2025qwen-image}, HunyuanImage-2.1~\cite{HunyuanImage-2.1}, and Seedream~4.0~\cite{gao2025seedream}, which demonstrating strong text-image alignment capabilities, thereby mitigating the need for extensive manual filtering during subsequent human evaluation. For the generation of each preference pair, we randomly select one of the three models to produce both the \textit{perfect} image and its corresponding \textit{perturbed} image, ensuring that each pair is generated consistently by the same model. As illustrated in Figure~\ref{fig:supp:statistics}, we present the distribution and proportions of preference pairs contributed by each model within our SpatialReward-Dataset.

\begin{figure}[t!]
    \centering
    \includegraphics[width=\linewidth]{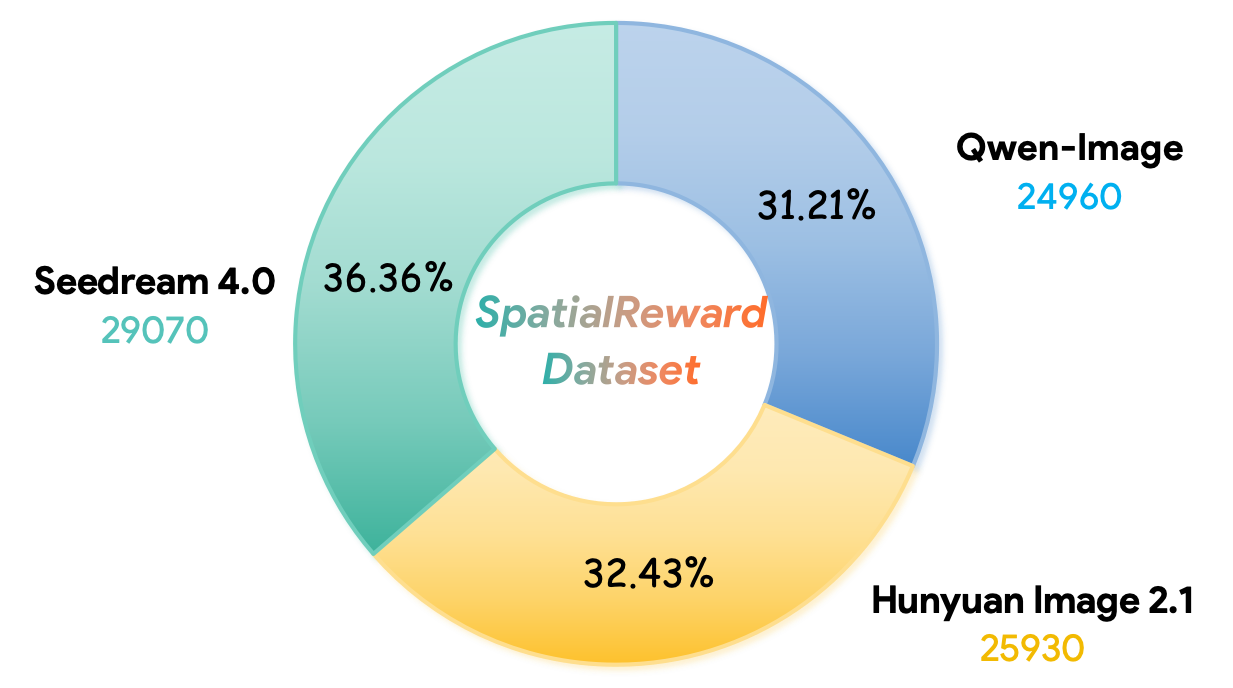}
    \vspace{-5pt}
    \caption{\textbf{Distribution statistics of SpatialReward-Dataset.}}
    \label{fig:supp:statistics}
    \vspace{-5pt}
\end{figure}

\begin{figure*}[t!]
    \centering
    \includegraphics[width=\linewidth]{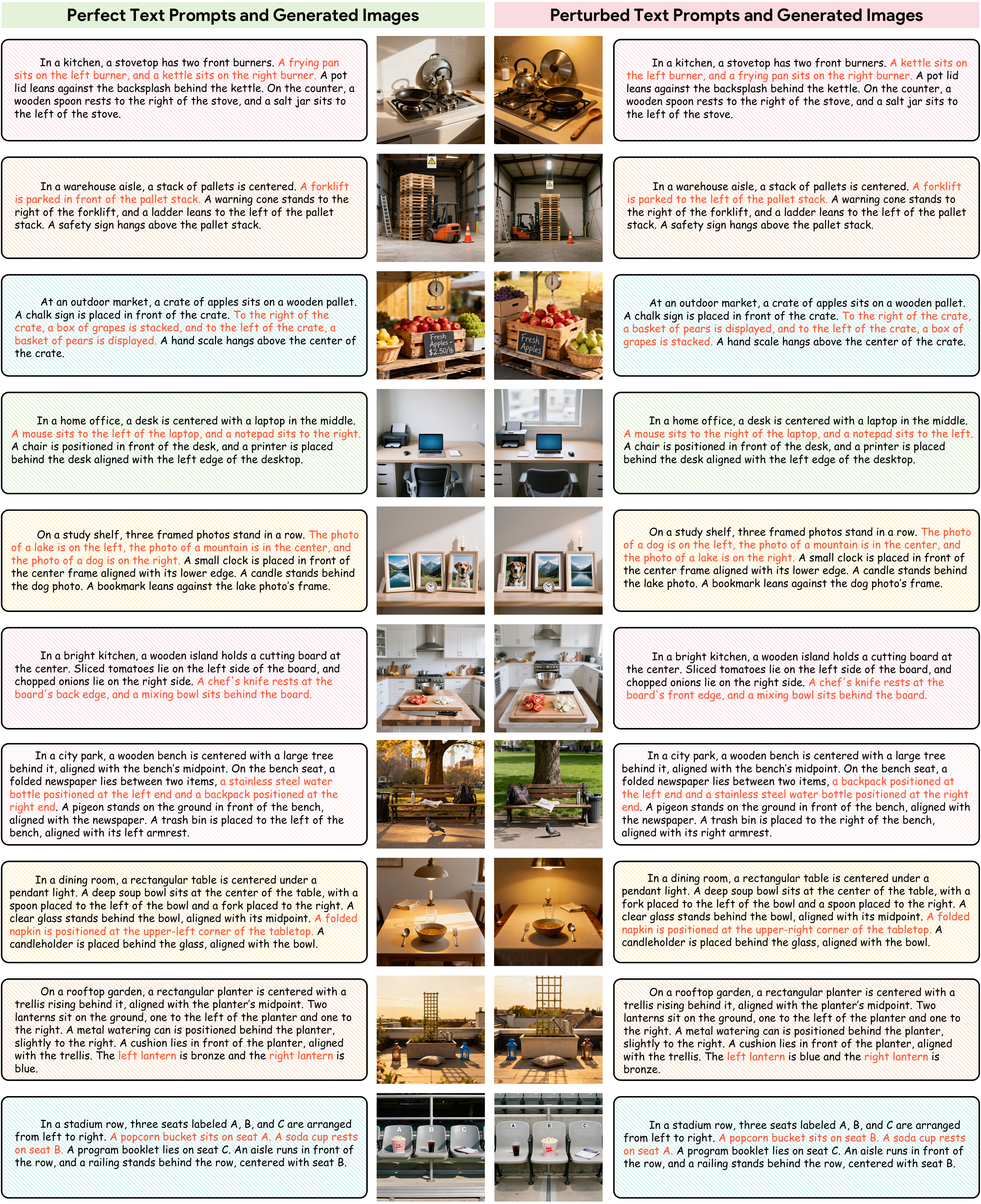}
    \vspace{-5pt}
    \caption{Visualization of the preference pairs (perfect images and perturbed images) in our SpatialReward-Dataset.}
    \label{fig:supp:more_data_pairs1}
    \vspace{-5pt}
\end{figure*}

\begin{figure*}[t!]
    \centering
    \includegraphics[width=\linewidth]{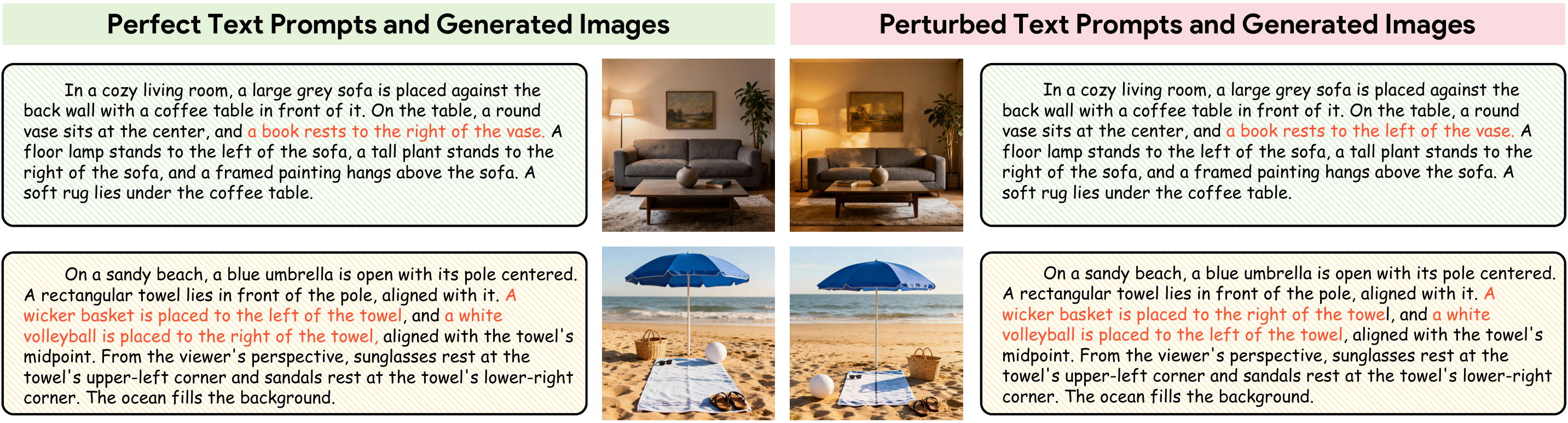}
    \vspace{-5pt}
    \caption{More visualizations of the preference pairs (perfect images and perturbed images) in our SpatialReward-Dataset.}
    \label{fig:supp:more_data_pairs2}
    \vspace{-5pt}
\end{figure*}

\subsection{Human Verifications}
After collecting the initial preference prompts and generating the corresponding preference pairs, we perform an additional round of human verification to ensure the reliability and overall quality of our \textsc{SpatialReward-Dataset}. Specifically, the verification follows two main principles:
\begin{enumerate}
\item Verification of perfect images.  We examine whether each \textit{perfect} image faithfully satisfies the complex spatial relationships across all objects in the prompt. If a perfect image contains clear violations of the specified spatial relations, the entire preference pair is discarded.

\item Verification of perturbed images.  We assess the spatial discrepancies between the \textit{perturbed} image and its corresponding \textit{perfect} image. In some cases, although the perturbed prompt differs from the original prompt, the resulting spatial relationships may remain nearly identical. If the perturbed image fails to exhibit the intended spatial deviation and instead shares the same spatial layout as the perfect image, the preference pair is removed.
\end{enumerate}

\section{Additional Experiment Results}

\subsection{Applying SpatialScore to Qwen-Image}
To further assess the effectiveness and robustness of our reward model \textsc{SpatialScore}, we perform RL fine-tuning on advanced image generation models Qwen-Image~\cite{wu2025qwen-image}. As shown in Table~\ref{tab:sup_qwen_image}, we observe consistent and significant improvements in spatial understanding compared to the base model Qwen-Image, similar to the comparisons on the Flux.1-dev~\cite{flux2024}. On the in-domain \textsc{SpatialScore} evaluation, our method improves from 6.74 to 8.25, demonstrating the effectiveness of RL training using \textsc{SpatialScore} as the reward model.  For other text-image alignment benchmarks, we select the spatial-aware sub-dimensions to assess spatial understanding, as discussed in the paper. As seen in Table~\ref{tab:sup_qwen_image}, our RL training yields consistent improvements across both short-prompt and long-prompt settings.

\begin{table}[t]
    \centering
    \caption{Quantitative evaluations on the Geneval benchmark. Our model is trained using SpatialScore as the reward model.}
    \vspace{-6pt}
    \label{tab:geneval_evaluation}
    \renewcommand{\arraystretch}{1.37}
    \resizebox{\linewidth}{!}{%
        \begin{tabular}{l|cccccc|c}
            \toprule
            Methods & \makecell{Single \\ object} & \makecell{Two \\ object} & Counting & Colors & Position & \makecell{Color \\ Attribute} & Overall \\
            % Methods & single\_object & two\_object & counting & colors & position & color\_attr & Overall \\
            \midrule
            Flux.1-dev & 0.99 & 0.81 & 0.70 & 0.76 & 0.19 & 0.45 & 0.65 \\
            \rowcolor{gray!10} 
            Ours & 1.00 & 0.92 & 0.88 & 0.79 & 0.37 & 0.66 & 0.78 \\
            \bottomrule
        \end{tabular}
    }
     \vspace{-5pt}
\end{table}

\subsection{Evaluations on Geneval Benchmark}
Despite the fact that Geneval as a reward model, often yields unreliable evaluations under visual challenges such as occlusion and exhibits limited generalization to long texts involving complex inter-object spatial relationships after RL training, we nonetheless utilize the Geneval benchmark to comprehensively assess the generalization of our model. Specifically, we conduct evaluations on the Geneval benchmark constructed by simple, fixed-template compositions. As shown in Table~\ref{tab:geneval_evaluation}, our model, which employs SpatialScore-guided RL training, achieved significant zero-shot improvements across all evaluated metrics.

\begin{table}[t] % 使用标准的 table 环境，确保只占用一栏宽度
  \centering
  \caption{Ablation study of SpatialScore backbone sizes on the reward evaluation benchmark. “1 Pert.” and “2–3 Pert.” denote subsets constructed by applying one or two–three spatial perturbations, respectively, to perfect prompts for perturbed prompts.}
  \vspace{-0.2cm}
  \renewcommand{\arraystretch}{1.2}
  % \setlength{\tabcolsep}{15pt} % 适当调整列间距，保持美观
  % 仅包含 Setting, Qwen2.5-VL-3B, Qwen2.5-VL-7B 三列
   \resizebox{0.466\textwidth}{!}{%
  \begin{tabular}{l c c c}
    \toprule
    \textbf{Setting\textbackslash backbone} & Qwen2.5-VL-3B & Qwen2.5-VL-7B & Qwen2.5-VL-32B\\
    \midrule
    1 Pert. & 0.861 & 0.939& 0.955 \\
    2–3 Pert. & 0.919 & 0.978 & 0.989\\
    \midrule
    \rowcolor{gray!10}
    Overall & 0.891 & 0.958& 0.973 \\
    \bottomrule
  \end{tabular}
  }
  \vspace{-6pt}
  \label{tab:qwen_scaling_subset_final}
\end{table}
\subsection{Ablations on Model Size} 
To investigate the generalization of our method, we train models with varying backbone sizes from the Qwen2.5-VL series~\cite{bai2025qwen2.5vl}. As depicted in Table~\ref{tab:qwen_scaling_subset_final}, the accuracy of pairwise preference prediction on our reward evaluation benchmark progressively increased from $0.891$ to $0.958$ and $0.973$ as the backbone size scaled from 3B to 7B and 32B, respectively. Furthermore, by referencing the comparisons in the paper, we observe that the performance of SpatialScore with 7B size exceeds that of Gemini 2.5 Pro~\cite{comanici2025gemini} and approaches the performance of the 32B variant. Consequently, considering the training efficiency for RL training, we finally selecte the 7B size configuration for all experiments.

\begin{table*}[t]
  \centering
  \caption{Detailed comparisons for the Qwen-Image family on SpatialScore, DPG-Bench, TIIF-Bench (short/long), and UnigenBench++ (short/long). * denotes RL-training with our SpatialScore as the reward model.
BR, AR, and RR denote basic relation, attribute+relation, and relation+reasoning.
Lay-2D/3D refer to layout-2D/3D. Unibench denotes UnigenBench++.}
  \vspace{-0.2cm}
  \renewcommand{\arraystretch}{1.36}
  \setlength{\tabcolsep}{6pt}
  \resizebox{\textwidth}{!}{%
  \begin{tabular}{l|*{12}{c}}
    \toprule
    \multirow{2}{*}{\textbf{Method}} &
    \multirow{2}{*}{\textbf{SpatialScore}} &
    \multicolumn{1}{c}{\textbf{DPG-bench}} &
    \multicolumn{3}{c}{\textbf{TIIF-bench-short}} &
    \multicolumn{3}{c}{\textbf{TIIF-bench-long}} &
    \multicolumn{2}{c}{\textbf{Unibench(short)}} &
    \multicolumn{2}{c}{\textbf{Unibench(long)}} \\
    \cmidrule(lr){3-3}\cmidrule(lr){4-6}\cmidrule(lr){7-9}\cmidrule(lr){10-11}\cmidrule(l){12-13}
    & & Relation-Spatial
      & BR & AR & RR
      & BR & AR & RR
      & Lay-2D & Lay-3D
      & Lay-2D & Lay-3D \\
    \midrule
    
  %   SD3.5-M
  %     & --   & --     
  %     & --   & --   & -- 
  %     & --   & --   & --
  %     & --   & --
  %     & --   & -- \\
  %     SD3.5-M
  %     & --   & --     
  %     & --   & --   & -- 
  %     & --   & --   & --
  %     & --   & --
  %     & --   & -- \\
  % \midrule
    Qwen-Image~\cite{wu2025qwen-image}
      & 6.74
      & 0.920
      & 0.865   & 0.756  & 0.704
      & 0.827 & 0.751 & 0.716
      & 0.864 & 0.852
      & 0.912  & 0.860 \\
    % Flow-GRPO*~\cite{liu2025flow-grpo}
    %   & 3.01
    %   & 0.742
    %   & 0.851 & 0.652  & 0.621
    %   & 0.577 & 0.510 & 0.482
    %   & 0.726 & 0.635 
    %   & 0.445 & 0.405   \\
    \rowcolor{gray!10} 
    Ours*
      & \textbf{8.25}
      & \textbf{0.958}
      & \textbf{0.899} & \textbf{0.792} & \textbf{0.791}
      & \textbf{0.871} & \textbf{0.801} & \textbf{0.780}
      & \textbf{0.908} & \textbf{0.917}
      & \textbf{0.926} &\textbf{0.893} \\
    \bottomrule
  \end{tabular}}
  \label{tab:sup_qwen_image}
\vspace{-0.2cm}
  
\end{table*}

\subsection{Understanding of Spatial Issues in T2I mdoels} The core issue is  the misalignment between training captions and complex inference prompts. Current T2I models are trained on MLLM-derived captions, which mainly describe the existence of multiple objects but lack complex constraints of spatial relationships between them. Thus, these models are prone to spatial errors for long, spatially complex prompts. Moreover, existing reward models mainly focus on aesthetics and semantic alignment, and Figure~\ref{fig:duck_placeholder}in our paper shows these models fail to correctly penalize spatial errors. We propose the first reward model designed for spatial understanding.
% \subsection{Ablations of Top-k Filtering on PickScore}
% To further validate the generalization of our Top-k filtering strategy, we utilize the widely adopted PickScore as our reward model for the online RL training using the Flux.1-dev~\cite{flux2024} as the base model. As illustrated in Figure~\ref{fig:supp:curve}, the incorporation of the Top-k filtering strategy yields a faster reward uplift compared to the baseline GRPO setup without this filtering, thereby enhancing training efficiency with the fewer number of function
% evaluations (NFEs).

\subsection{More Clarifications on NFE}
In the paper, we present comparisons on the number of function evaluations (NFEs)~\cite{lu2022dpm} for ablations of Top-$k$ filtering. Here, we provide a detailed explanation of its calculation. Specifically, NFE refers to the number of forward passes of the policy model exclusively for the computation of the policy ratio $r(\theta)$ during the training stage, and not the sampling stage. Given a sampling group size of 24 per prompt and 6 denoising steps, the NFEs differ based on the setup. For the original GRPO setup without top-$k$ filtering, the requirement is $24 \times 6$ NFEs per prompt at each training step for the training stage. For our GRPO setup adding top-$k$ filtering, we select a total of $2\times k$ samples (comprising top-$k$ and bottom-$k$) after the sampling stage. We then use these filtered samples for the training stage, consequently requiring $2\times k \times 6$ NFEs per prompt at each training step.

\subsection{More Visual Demos}
As shown in Figure~\ref{fig:supp:more_qua_results}, we provide more qualitative comparisons between the base model Flux.1-dev~\cite{flux2024} and the Flux variant trained with Flow-GRPO on the Geneval. Our RL-trained model with \textsc{SpatialScore} demonstrates improved spatial understanding, generating images that more accurately reflect the complex spatial relationships between multiple objects as described in the prompts. In contrast, the Flux.1-dev variant, trained with Flow-GRPO on Geneval, exhibits limited generalization for complex spatial relationships, even losing part of the base model’s ability to follow long prompts. As shown in Figure~\ref{fig:supp:more_qua_results}, this includes missing key objects, such as the candles in Example (1) and the soap and toothbrush in Example (2). Furthermore, due to its reliance on rule-based rewards from object detectors, the Flux variant guided by the Geneval frequently generates images with visually implausible artifacts, such as the napkin floating in mid-air, as seen in Example (5).

% There are some sentences to fill this paper. There are some sentences to fill this paper. There are some sentences to fill this paper. There are some sentences to fill this paper. There are some sentences to fill this paper. There are some sentences to fill this paper. There are some sentences to fill this paper. There are some sentences to fill this paper. There are some sentences to fill this paper. There are some sentences to fill this paper. There are some sentences to fill this paper. There are some sentences to fill this paper. There are some sentences to fill this paper. There are some sentences to fill this paper. There are some sentences to fill this paper. There are some sentences to fill this paper. There are some sentences to fill this paper. There are some sentences to fill this paper. There are some sentences to fill this paper. There are some sentences to fill this paper. There are some sentences to fill this paper. There are some sentences to fill this paper. 

% \begin{figure}[t!]
%     \centering
%     % \includegraphics[width=\linewidth]{Figures/fig_supp_circle_v2.png}
%     \includegraphics[width=0.886\linewidth]{Figures/supp_wandb_log2.pdf}
%     \vspace{-5pt}
%     \caption{Reward training curves for ablation study of top-$k$ filtering When using PickScore as the reward model.}
%     \label{fig:supp:curve}
%     \vspace{-5pt}
% \end{figure}

\section{Limitations and Future Works}
Although we have validated the enhancement of spatial understanding through reward modeling at the image level, the integration of spatial understanding with temporal dynamics has not been fully explored, particularly with regard to video generation. Video generation requires models to not only comprehend static spatial relationships but also to account for dynamic changes over time. For instance, a model may need to move object A to the left of object B, then place object C to the right of object B, and subsequently swap  the positions of objects A and C. Therefore, an essential challenge for future work will be to effectively extend reward modeling to enhance the spatial understanding of video generation models. This direction is especially important for sim-to-real embodied simulation scenarios, where generating temporally consistent and spatially accurate video sequences is critical for bridging the gap between simulated environments and real-world dynamics.

\begin{figure*}[t!]
    \centering
    \includegraphics[width=0.95\linewidth]{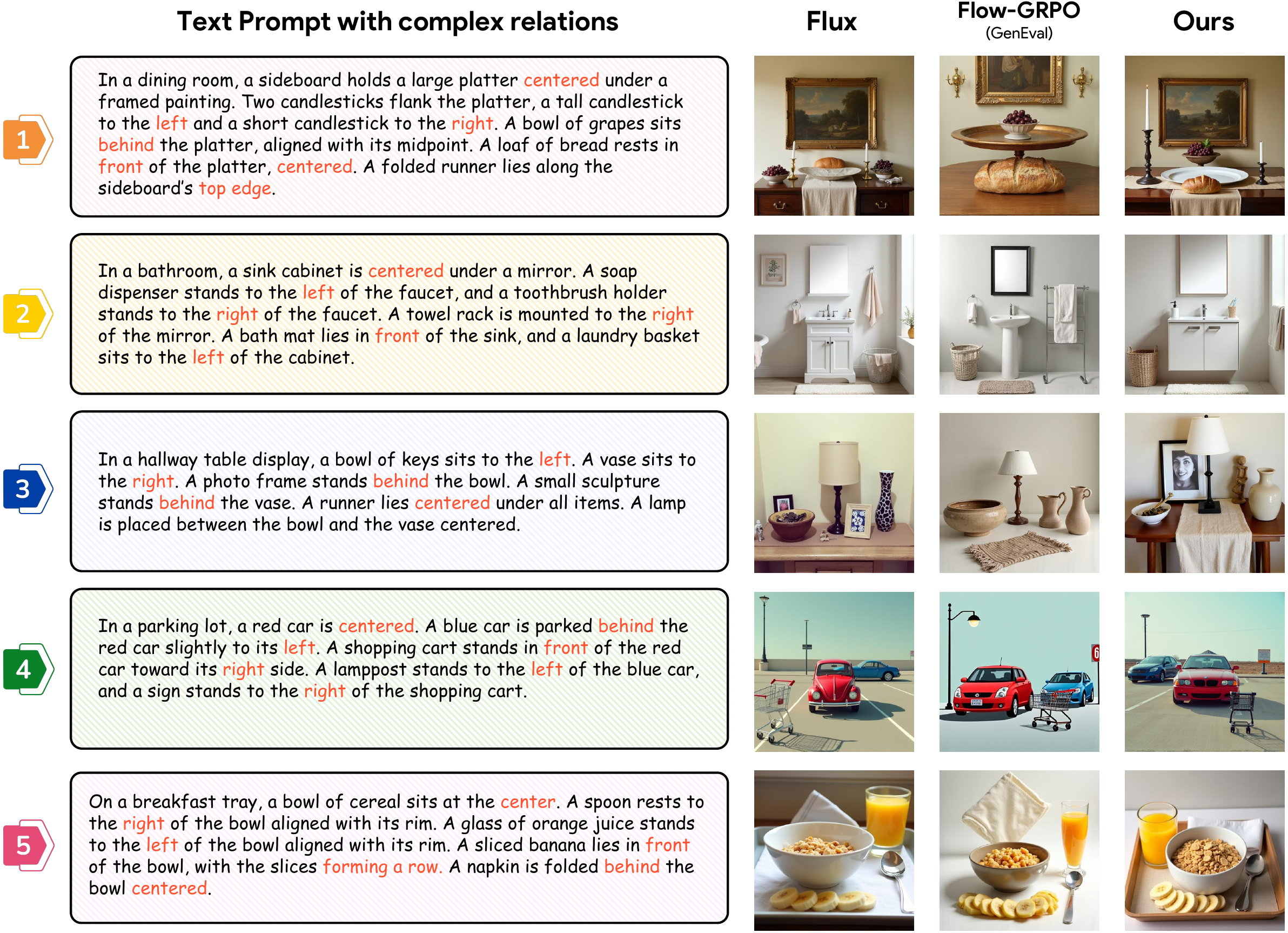}
    \vspace{-5pt}
    \caption{More qualitative comparisons of the long prompts with complex spatial relationships across objects.}
    \label{fig:supp:more_qua_results}
    \vspace{-5pt}
\end{figure*}

% \section{Rationale}
% \label{sec:rationale}
% % 
% Having the supplementary compiled together with the main paper means that:
% % 
% \begin{itemize}
% \item The supplementary can back-reference sections of the main paper, for example, we can refer to \cref{sec:intro};
% \item The main paper can forward reference sub-sections within the supplementary explicitly (e.g. referring to a particular experiment); 
% \item When submitted to arXiv, the supplementary will already included at the end of the paper.
% \end{itemize}
% % 
% To split the supplementary pages from the main paper, you can use \href{https://support.apple.com/en-ca/guide/preview/prvw11793/mac#:~:text=Delete%20a%20page%20from%20a,or%20choose%20Edit%20%3E%20Delete).}{Preview (on macOS)}, \href{https://www.adobe.com/acrobat/how-to/delete-pages-from-pdf.html#:~:text=Choose%20%E2%80%9CTools%E2%80%9D%20%3E%20%E2%80%9COrganize,or%20pages%20from%20the%20file.}{Adobe Acrobat} (on all OSs), as well as \href{https://superuser.com/questions/517986/is-it-possible-to-delete-some-pages-of-a-pdf-document}{command line tools}.

% % \clearpage
% {
%     \small
%     \bibliographystyle{ieeenat_fullname}
%     \bibliography{main}
% }
\end{document}